\documentclass[letterpaper]{article} 
\usepackage{aaai2027}  
\usepackage[hyphens]{url}  
\usepackage{graphicx} 
\usepackage{natbib}  
\usepackage{caption} 
\usepackage{algorithm}

\usepackage{algpseudocode}
\usepackage{siunitx}
\graphicspath{ {./imgs/} }
\usepackage{amsmath}
\usepackage{amssymb}

\usepackage{newfloat}
\usepackage{listings}
\DeclareCaptionStyle{ruled}{labelfont=normalfont,labelsep=colon,strut=off} 
\floatstyle{ruled}
\newfloat{listing}{tb}{lst}{}
\floatname{listing}{Listing}

\usepackage{booktabs}

\title{How Edge of Stability Hinders SCAFFOLD in Federated Optimization}
\nocopyright  

\author {
    Anant Khandelwal\textsuperscript{\rm 1,}\corresponding,
    Michael Crawshaw\textsuperscript{\rm 2,3,}\corresponding,
    Mingrui Liu\textsuperscript{\rm 2}
}
\affiliations {
    \textsuperscript{\rm 1}Georgia Institute of Technology, College of Computing\\
    \textsuperscript{\rm 2}George Mason University, Department of Computer Science\\
    \textsuperscript{\rm 3}Flatiron Institute, Center for Computational Mathematics\\
    \texttt{akhandelwal79@gatech.edu}, \texttt{michael.crawshaw.4@gmail.com}, \texttt{mingruil@gmu.edu}
}

\begin{document}

\maketitle

\begin{abstract}
In federated learning, it is well known that heterogeneous data can (in theory) slow down optimization, and much effort has been directed at designing optimization algorithms that are unaffected by data heterogeneity, such as the SCAFFOLD algorithm. Yet, despite strong theoretical guarantees, SCAFFOLD does not usually outperform the much simpler FedAvg in practice. In this work, we propose that this gap is due to the presence of Edge of Stability (EoS) and progressive sharpening in federated optimization, supported by extensive empirical probing. First, we find that EoS-like dynamics occur with both FedAvg and SCAFFOLD under a variety of architectures and hyperparameters. We observe that the equilibrium value of the sharpness is inversely proportional to the learning rate (as in GD), and interestingly, the degree of data heterogeneity (but not the number of local steps) also affects the equilibrium value. Most importantly, we observe that SCAFFOLD's ability to estimate the gradient of the global objective is severely degraded at the EoS, as measured by the correlation between sharpness and SCAFFOLD's error in estimating the global gradient along the optimization trajectory. This suggests a mechanism for SCAFFOLD's lackluster performance in deep learning: with high sharpness at the EoS, SCAFFOLD cannot reliably estimate the global gradient.
\end{abstract}

\section{Introduction}
Federated learning is a paradigm for machine learning in which many separate machines
(often user devices) communicate over a network to collaboratively train a machine
learning model, which leverages hardware and data from many devices while preserving
privacy \citep{mcmahan2017communication, kairouz2021advances, wang2021field}. However,
the federated optimization process can be slowed down when user data exhibits large
heterogeneity across devices, and indeed for standard algorithms like FedAvg
\citep{mcmahan2017communication}, the theoretical convergence rate slows as user data
becomes more heterogeneous \citep{woodworth2020minibatch, koloskova2020unified}.


The SCAFFOLD algorithm \citep{karimireddy2020scaffold} was introduced to eliminate this
slowdown from heterogeneous data, by using control variates to estimate the gradient of
the global objective via gradients from previous communication rounds. In theory,
SCAFFOLD enjoys a convergence rate that is independent of data heterogeneity, which
suggests that SCAFFOLD should be a strict improvement over FedAvg. However, it has been
observed that in deep learning, SCAFFOLD's performance is lacking compared to the
theoretical bounds \citep{yu2022tct, li2023effectiveness, baumgart2024not}.
Indeed, in practice FedAvg is "unreasonably effective" \cite{wang2024on,
patel2023on} and remains preferred over SCAFFOLD. This discrepancy between
theory and practice motivates our central question:

\begin{center}
\textbf{\textit{
Why does FedAvg match or outperform SCAFFOLD in practice, despite theory suggesting otherwise?
}}
\end{center}



\citet{yu2022tct} observe
empirically that SCAFFOLD works well when training linear models, but doesn't
outperform FedAvg under the nonconvexity of training deep networks, and conclude
that the existing "black-box" theory is insufficient for capturing the
difficulty of training deep networks. Also, \citet{wang2024on} theoretically
investigate the success of FedAvg through alternative assumptions on the
objective heterogeneity. However, we are not aware of any work that proposes a
mechanism which explains why SCAFFOLD's optimization performance does not
improve over FedAvg when training deep networks, despite theoretical guarantees.

In this paper, we conjecture that SCAFFOLD's poor performance is due to the phenomena of
progressive sharpening and Edge of Stability (EoS) \citep{cohen2021gradient}, in which
the objective's Hessian norm (sharpness) along the trajectory of Gradient Descent
(GD) increases until equilibrating around $2/\eta$ (where $\eta$ is the
learning rate). Progressive sharpening tends to increase the sharpness up
to a threshold that depends on $\eta$, while SCAFFOLD's theoretical
guarantees only apply in settings where the sharpness is a priori known to be bounded
and the learning rate/number of local steps can be chosen using this bound.


We perform an extensive empirical probe to investigate our conjecture:
\begin{itemize}
    \item We observe in Section \ref{sec:exp_eos} that, for image classification tasks,
        both \textbf{FedAvg and SCAFFOLD exhibit progressive sharpening and EoS}
        in terms of local objectives and global objectives. The occurrence is
        consistent across network architectures and optimizer hyperparameters.
    \item In Section \ref{sec:exp_threshold}, we find that the equilibrium sharpness
        for both FedAvg and SCAFFOLD is inversely proportional to the learning rate
        (similarly to GD), but also depends to a smaller degree on the severity of
        heterogeneity. Interestingly, \textbf{the number of local steps does not
     influence the equilibrium sharpness}, which eliminates the
        possibility of satisfying theoretical conditions for SCAFFOLD's
        convergence when the number of local steps is large.
    \item We find in Section \ref{sec:exp_misalign} that \textbf{SCAFFOLD's
        ability to mitigate data heterogeneity is significantly degraded at the
        EoS}. We use a new metric, called the \textit{update misalignment}, which measures the difference
        between an algorithm's update compared to the gradient of the global
        objective. We find that SCAFFOLD's update misalignment throughout
        training is strongly correlated with sharpness, supporting our
        conjectured connection between EoS and SCAFFOLD's lackluster
        performance. Essentially, SCAFFOLD does not reliably estimate the global
        gradient when sharpness is high, while FedAvg's update misalignment is
        relatively unaffected by high sharpness.
\end{itemize}

Overall, our results suggest an explanation for the fact that SCAFFOLD does not improve
over FedAvg in deep learning: both algorithms experience progressive sharpening and EoS,
but under the resulting high sharpness, SCAFFOLD cannot reliably estimate the global
gradient.

The rest of this paper is structured as follows. We discuss related work in Section
\ref{sec:related_work}, then review background and establish preliminaries in Section
\ref{sec:preliminaries}. Our main results are in Section \ref{sec:experiments}, where we
perform an experimental probe of EoS in federated learning, and we conclude with Section
\ref{sec:conclusion}.




\section{Related Work} \label{sec:related_work}

\subsection{Federated Learning and SCAFFOLD}
Federated learning \citep{mcmahan2017communication} is a paradigm of machine learning in
which a model is trained collaboratively across multiple machines, often with differing
datasets, that communicate over a network. The practice has come into increasing use in
tandem with the shift towards mobile devices, IoT sensors, and distributed systems. \citep{LIU2024128019}. The de facto
standard optimization algorithm for federated learning is FedAvg a.k.a. Local SGD, which
is a parallelized version of SGD in which multiple update steps are executed locally by
each worker machine between synchronizations across all workers. The convergence
behavior of FedAvg is well-studied \citep{stich2019local, haddadpour2019convergence,
woodworth2020minibatch, khaled2020tighter, koloskova2020unified, glasgow2022sharp}, and
in particular it is known that convergence can slow down as worker data becomes more
heterogeneous.

SCAFFOLD \citep{karimireddy2020scaffold} was introduced to alleviate this problem: it
provably solves various federated optimization problems (e.g. smooth, non-convex,
stochastic, heterogeneous) with a rate that does not depend on data heterogeneity. This
is achieved by the use of ``gradient corrections", in which updates to local models use
an approximation of the gradient of the global objective, rather than the gradient of
the local objective as in FedAvg (see pseudocode in Algorithm \ref{alg:scaffold}). These
gradient corrections are constructed using historical values of the gradient, and
SCAFFOLD rests on the premise that gradients encountered in previous communication
rounds will be a good estimate of gradients in the current round.

 However, multiple works \citep{yu2022tct,
li2023effectiveness, baumgart2024not} noted that SCAFFOLD can struggle when training
deep neural networks. Both \citet{yu2022tct} and \citet{li2023effectiveness} proposed
variations of the SCAFFOLD algorithm that only apply gradient corrections to the final
layer of a neural network, which empirically improve upon SCAFFOLD. In an empirical comparison of federated optimization algorithms,
\citet{baumgart2024not} found that SCAFFOLD fails in many cases, particularly with high
heterogeneity, and fails catastrophically without gradient clipping.

On the theoretical side, \citet{cheng2024momentum} found that the use of momentum
provably increases the rate of convergence for SCAFFOLD, and SCAFFOLD with momentum
outperforms both FedAvg and SCAFFOLD. More recent analyses \citep{luo2025revisiting,
mangold2025scaffoldstochasticgradientsnew} have focused on provable convergence
guarantees under various convexity, smoothness, and heterogeneity assumptions
\cite{luo2025revisiting} and fine-grained characterization of the limiting distribution
of SCAFFOLD's iterates \cite{mangold2025scaffoldstochasticgradientsnew}.


FedAvg, on the contrary, seems to perform well under heterogeneity, despite its lack of
theoretical guarantees implying so. \citet{wang2024on} described this as the
"unreasonable effectiveness" of FedAvg, and propose to explain the theory-practice gap
with new theoretical conditions capturing heterogeneity near optima. Follow-up work
\cite{patel2023on} provided lower bounds which they argue rules out the heterogeneity
conditions proposed by \citet{wang2024on} as a sufficient explanation of this
unreasonable effectiveness. A parallel line of work \citep{patel2022towards,
patel2024limits, patel2026revisiting} investigates whether the success of FedAvg can be
explained through higher-order smoothness and heterogeneity assumptions.



\subsection{Edge of Stability}
The Edge of Stability (EoS) \citep{cohen2021gradient} is a phenomenon where the maximum
eigenvalue of the Hessian (a.k.a. sharpness) gradually increases along the trajectory of
gradient descent when training neural networks, until eventually equilibrating around
the threshold $2/\eta$, at which point the training loss exhibits oscillations in the
short term while generally decreasing in the long term. The apparent dependence of the
sharpness on the learning rate is a departure from smooth optimization theory, in which
it is assumed that the sharpness is globally bounded, and the learning rate can be
chosen in terms of this bound. The EoS phenomenon was first noticed for GD with a
variety of architectures, hyperparameters, and datasets \citep{cohen2021gradient}, then
extended to adaptive optimizers \citep{cohen2023adaptive} and clarified with the Central Flows framework
\citep{cohen2025understanding}.



There are many works that analyze or propose explanations for EoS.
\citet{arora2022understandinggradientdescentedge} showed that GD and normalized GD
follow a flow on the manifold of minimizers in the direction that decreases sharpness,
under certain general conditions on the loss. Other works have studied surrogate models
of deep learning, such as 4-layer scalar networks \citep{zhu2023understanding},
one-neuron neural networks \citep{chen2023beyond}, diagonal linear networks
\citep{even2023s}, and a two-parameter model of ReLU networks \citep{ahn2023learning}.
Notably, \citep{damian2023selfstabilization} proved a self-stabilization effect for GD
on general objectives with a progressive sharpening property, by using a third-order
Taylor series analysis that demonstrates how regions of sharpness exceeding $2/\eta$
tend to push the trajectory back into lower sharpness regions.



Our aim in this work is not to explain EoS, but rather to establish its occurrence and
consequences in federated optimization. For a thorough review of the literature on this
topic, see \citet{cohen2025understanding}.

\section{Preliminaries} \label{sec:preliminaries}
We consider the federated optimization problem, consisting of $M$ local objectives $f_i:
\mathbb{R}^d \rightarrow \mathbb{R}$, where the goal is to minimize the global objective
$f(w) = \frac{1}{M} \sum_{i=1}^M f_i(w)$. Each local objective corresponds to the
loss for a single client in a collaborative federation, so the
differences in the data of clients can induce differences in local objectives, which is
called data heterogeneity. Large levels of
heterogeneity can slow down optimization for FedAvg \citep{woodworth2020minibatch},
which is the de facto standard algorithm for federated optimization. See Algorithm
\ref{alg:fedavg} for the pseudocode of FedAvg.

\paragraph{SCAFFOLD} SCAFFOLD \citep{karimireddy2020scaffold} (shown in Algorithm
\ref{alg:scaffold}) was introduced to circumvent slow optimization under heterogeneous
objectives. Under the assumption that each $f_i$ is $L$-smooth (that is, $\|\nabla^2
f_i(w)\| \leq L$ for every $w$), and with an appropriate choice of hyperparameters,
SCAFFOLD can solve the federated optimization problem with a convergence rate
independent of heterogeneity, which is achieved through the use of \textit{control
variates} to estimate the gradient of the global objective.

Let $w_i^{t,k}$ denote the local model of client $i$ during the
$k$-th local step of communication round $t$, and let $g_i^{t,k}$ denote the gradient
computed at $w_i^{t,k}$. SCAFFOLD constructs a control variate $c_i^t = \frac{1}{K}
\sum_{k=0}^{K-1} g_i^{t-1,k}$, i.e. the average of the gradients of $f_i$ computed
during communication round $t-1$, then computes an average control variate $c^t =
\frac{1}{M} \sum_{i=1}^M c_i^t$. During local step $k$ of round $t$, instead of updating
each local model in the direction $g_i^{t,k}$ (as in FedAvg), SCAFFOLD updates in the
direction $g_i^{t,k} - c_i^t + c^t$. This update direction
approximates the global gradient; the key is that gradients from the previous round are
representative of gradients in the current round, as long as each $f_i$ is $L$-smooth
and we choose the stepsize $\eta \leq \mathcal{O}(1/KL)$ (see e.g. Theorem III of
\cite{karimireddy2020scaffold}).

\paragraph{SCAFFOLD and Sharpness}
However, SCAFFOLD's assumption of $L$-smoothness clashes with the empirically observed
phenomena of progressive sharpening and Edge of Stability (EoS) phenomenon in deep
learning optimization. Progressive sharpening is the gradual increase in the loss's
Hessian norm (the sharpness) along the trajectory of gradient descent, and EoS is the
equilibration of the Hessian norm around the threshold $2/\eta$ (where $\eta$ is the
learning rate of GD), which is accompanied by non-monotonic behavior of the loss over
time. Indeed, such a large sharpness precludes the guarantee of monotonic loss decrease
ensured by the classical learning rate choice $\eta < 2/L$ when the loss is $L$-smooth.
Essentially, $L$-smoothness assumes that we can choose the stepsize in terms of a fixed
smoothness constant $L$, while EoS suggests the reverse causal relationship, that the
smoothness constant along the training trajectory adapts to the chosen stepsize.

SCAFFOLD argues that the update direction $\delta_{r,k}^i := \nabla F_i(x_{r,k}^i) -
c_r^i + c_r$ of SCAFFOLD approximates the global gradient at the current local model,
that is, $\delta_{r,k}^i \approx \nabla F(x_{r,k}^i)$. This is justified by the fact
that for each $j$,
\begin{equation} \label{eq:cv_approx}
    c_r^j \approx \nabla F_j(x_{r,k}^i),
\end{equation}
so that the update direction satisfies
\begin{align}
    \delta_{r,k}^i &\approx \nabla F_i(x_{r,k}^i) - \nabla F_i(x_{r,k}^i) + \frac{1}{N} \sum_{j=1}^N \nabla F_j(x_{r,k}^i)\\ &= \nabla F(x_{r,k}^i).
\end{align}
This argument hinges on Equation \ref{eq:cv_approx}, which is justified based on the
assumption that each $F_j$ is $L$-smooth: we can bound the error of the approximation in
Equation \ref{eq:cv_approx} by

\begin{align}
    \|\nabla F_j(x_{r,k}^i) - c_r^j\| &= \left\| \frac{1}{K} \sum_{\ell=1}^K (\nabla F_j(x_{r,k}^i) - \nabla F_j(x_{r-1,\ell}^i)) \right\| \\ &\leq \frac{1}{K} \sum_{\ell=1}^K \left\| \nabla F_j(x_{r,k}^i) - \nabla F_j(x_{r-1,\ell}^i) \right\| \\
    &\leq \frac{L}{K} \sum_{\ell=1}^K \left\| x_{r,k}^i - x_{r-1,\ell}^i \right\|.
\end{align}

Notice that the distance between parameters of consecutive rounds $\left\| x_{r,k}^i -
x_{r-1,\ell}^i \right\|$ is proportional to both $\eta$ and the number of local steps
$K$, so in the end the error $\|\nabla F_j(x_{r,k}^k) - c_r^j\|$ is proportional to
$\eta KL$. Accordingly, the convergence proof of SCAFFOLD requires the condition $\eta
\leq \mathcal{O}(1/KL)$ to control this error (see Theorem III of
\citep{karimireddy2020scaffold}). Equivalently, the sharpness should be smaller than
$\mathcal{O}(1/\eta K)$, which goes to zero as $K$ grows.
\begin{figure*}
    \centering
    \includegraphics[width=1.0\linewidth]{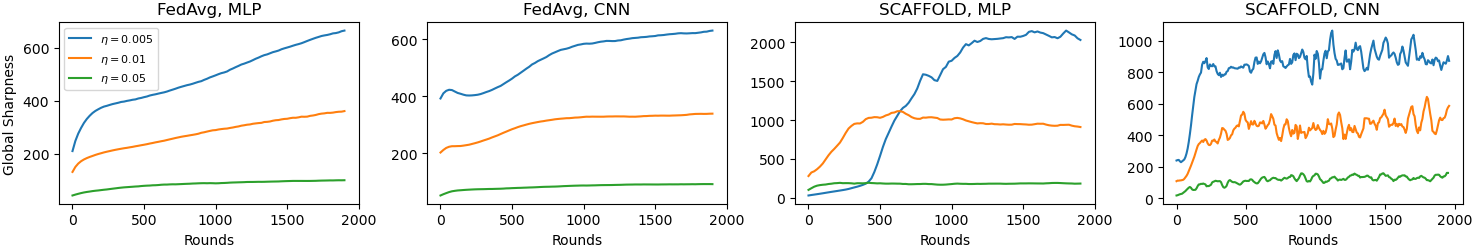}
    \includegraphics[width=1.0\linewidth]{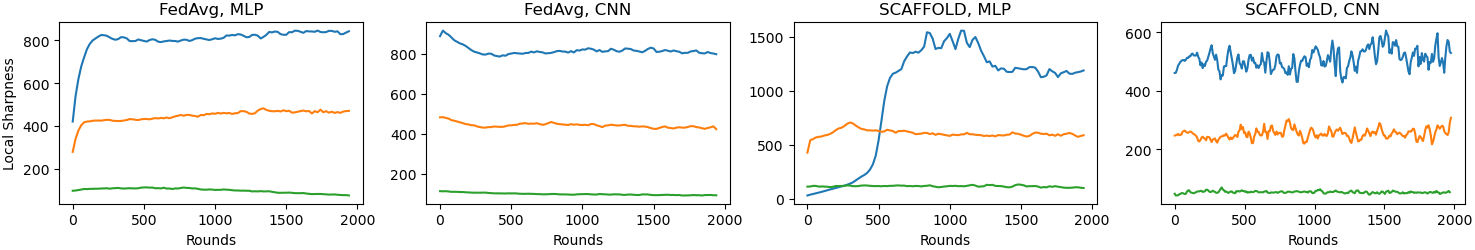}
    \includegraphics[width=1.0\linewidth]{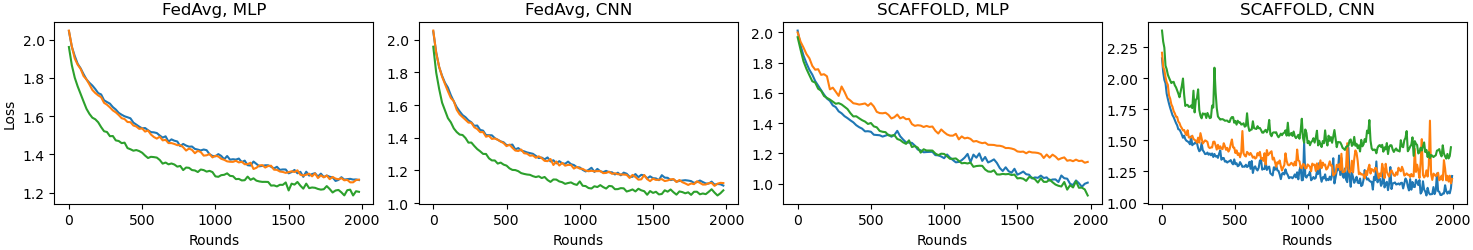}
    \caption{Sharpness and loss trajectories for FedAvg (left) and SCAFFOLD (right) on CIFAR-10 with MLPs and CNNs (three layers). We measure sharpness of the global objective (top), sharpness of the local objectives (middle), and training loss (bottom). Both SCAFFOLD and FedAvg experience progressive sharpening in terms of the global sharpness and the local sharpness.}
    \label{fig:cif-sharp}
\end{figure*}

If the sharpness hovers around a value that grows depending on our hyperparameters, then
it may not be possible to satisfy SCAFFOLD's requirement of a sharpness smaller than
$\mathcal{O}(1/\eta K)$ when training deep neural networks, which would hinder
SCAFFOLD's ability to estimate the global gradient. Crucially, we observe empirically
that as the communication interval $K$ grows, \textit{the equilibrium sharpness level
does not significantly change} (see Section \ref{sec:exp_threshold}). This means that
the required bound on the sharpness in order for SCAFFOLD to function gets smaller and
smaller as $K$ grows, while the actual sharpness along the trajectory remains the same.

Accordingly, we conjecture that SCAFFOLD's lackluster performance in deep learning is
closely tied to the occurrence of EoS: when sharpness is high, SCAFFOLD cannot
accurately estimate the global gradient.

\section{Experiments} \label{sec:experiments}
In this section, we perform an empirical investigation of EoS for FedAvg and SCAFFOLD.
In Section \ref{sec:exp_eos}, we establish that both algorithms exhibit EoS-like
dynamics, and in Section \ref{sec:exp_threshold} we investigate how the equilibrium
sharpness value is affected by learning rate, data heterogeneity, and communication
interval. Section \ref{sec:exp_misalign} corroborates our conjectured connection between
EoS and SCAFFOLD's inability to outperform FedAvg, by showing that SCAFFOLD's estimate
of the global gradient becomes less accurate at the EoS.


All experiments are run on image classification tasks of CIFAR-10 \citep{krizhevsky2009learning}, MNIST \citep{lecun1998gradient}, FashionMNIST \citep{xiao2017fashion} as simple CNN and MLP tasks consistent with previous works. 
using fully connected networks or simple CNNs with three to seven layers (for full architectures see \ref{app:model-architectures}. Following the
convention in many works on EoS \citep{cohen2021gradient, cohen2023adaptive,
cohen2025understanding}, we use full-batch gradients, which limits us to smaller
architectures due to hardware constraints. Unless otherwise noted, we use a sample of
5000 images from each dataset, split among eight clients using a common protocol to
induce heterogeneity among clients \citep{karimireddy2020scaffold}. The heterogeneity
protocol is defined in Appendix \ref{app:het}, and is parameterized by a scalar $h \in
[0, 1]$, with $h=0$ inducing i.i.d. data and $h=1$ inducing maximal data heterogeneity.
We refer to $h$ simply as the heterogeneity. Unless otherwise noted, we use a
learning rate of $\eta =0.01$, heterogeneity of $h=0.9$ and a communication interval of $K = 32$. All experiments were
run on a node of 8 NVIDIA A100 GPUs. Training runs are single trajectories unless otherwise stated. 


\begin{figure*}[t]
    \centering
    \begin{minipage}[t]{0.9\linewidth}
    \begin{center}
        \includegraphics[width=\linewidth]{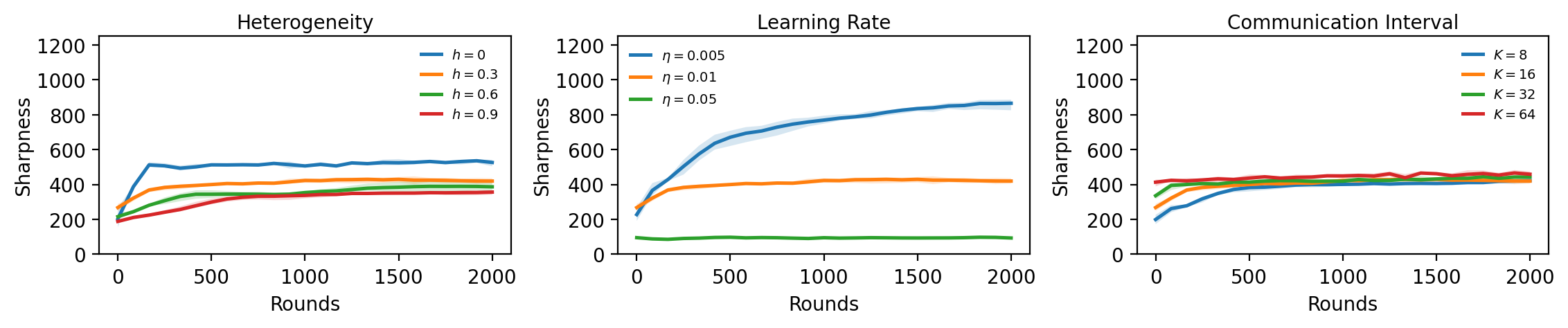}
        \small (a) FedAvg\\[0em]
    \end{center}
    \end{minipage}
    \begin{minipage}[t]{0.9\linewidth}
    \begin{center}
        \includegraphics[width=\linewidth]{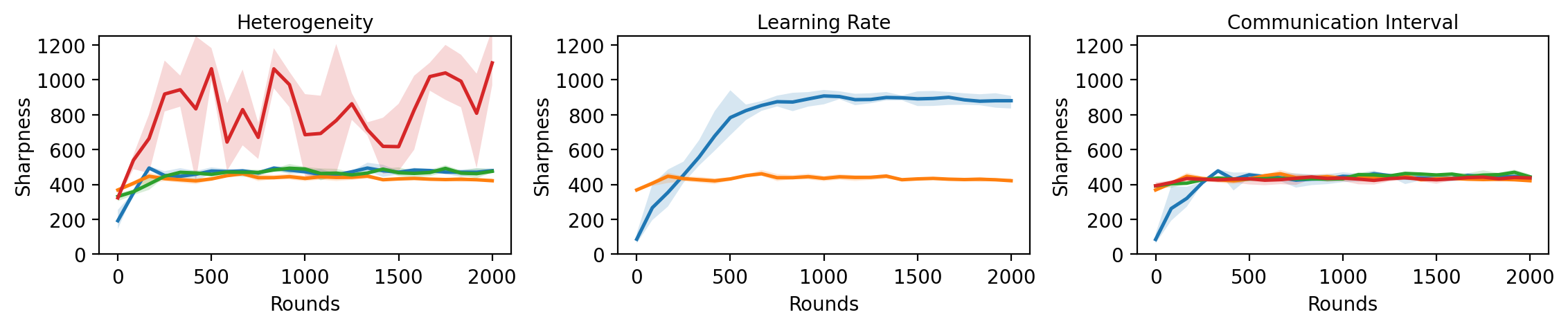}
        \small (b) SCAFFOLD\\[0em]
    \end{center}
    \end{minipage}
    \caption{Sharpness of local objectives with varying heterogeneity, learning rate,
    and communication interval (three-layer CNN, CIFAR-10 dataset), averaged over three
    random seeds. Error bars denote min and max values among 3 samples. Note that SCAFFOLD training with a learning rate
    of 0.05 diverged, so we omit the corresponding curve. Federated algorithms only have
    a clear inverse relation to learning rate, with minimal relation to heterogeneity or
    communication interval except for SCAFFOLD in the case of very high heterogeneity.}
    \label{fig:sharp-char}
\end{figure*}
\subsection{Progressive Sharpening Occurs In Federated Settings with High Heterogeneity} \label{sec:exp_eos}

To measure sharpness during federated learning trajectories, we define both the \textit{global sharpness} and the \textit{local sharpness}. The global sharpness tracks the sharpness of the averaged model on the combined dataset at the end of a round. The local sharpness is the average sharpness of the local models on the local datasets during each local step of the round. For formal definitions, see Appendix \ref{app:sharp}. In Sections \ref{sec:exp_threshold} and \ref{sec:exp_misalign}, we primarily use local sharpness.

Figure \ref{fig:cif-sharp} shows that with both FedAvg and SCAFFOLD, we observe
progressive sharpening and EoS-like behavior with both architectures and several
learning rates. Interestingly, the general pattern of gradual increase followed by
equilibration is more pronounced for SCAFFOLD, although both algorithms generally
exhibit an increase in at least one or either local sharpness or global sharpness.
Similarly as in the case of GD, we see that non-monotonic behavior of the loss is
associated with equilibration of the sharpness at EoS; for example, when SCAFFOLD trains
an MLP with $\eta = 0.005$, the loss is essentially monotonic for the first 500 steps,
at which point the local sharpness reaches its equilibrium value (about 1250) and the
loss becomes non-monotonic. In general, SCAFFOLD exhibits more erratic oscillations of
both the sharpness and the loss.


We similarly study the sharpness behavior for MNIST and FashionMNIST, shown in Figures
\ref{fig:fashionmnist}, \ref{fig:mnist-sharp} of Appendix \ref{app:extra_exp}. In these
easier settings, we notice that for both algorithms the sharpness declines significantly
before ever reaching an equilibrium value, and in many such cases the loss is monotonic
or eventually monotonic. This is consistent with GD in the single-machine setting
\citep{cohen2021gradient}. Importantly, these settings provide a testbed to study FedAvg
and SCAFFOLD when training deep networks without EoS; we will leverage this testbed to
probe SCAFFOLD in Section \ref{sec:exp_misalign}.


\subsection{What is the Equilibrium Sharpness Value?} \label{sec:exp_threshold}
Looking at Figure \ref{fig:cif-sharp}, we notice that that sharpnesses don't appear to
oscillate around the usual $2/\eta$ threshold from the single-machine case. In fact, in
some cases the global sharpness continues increasing while the local sharpness
equilibrates. It is natural then to ask, around which value does the sharpness tend to
oscillate?

To answer this question, we compare the equilibrium sharpness values under various
choices of the learning rate, data heterogeneity $h$, and communication interval $K$,
shown in Figure \ref{fig:sharp-char}. Here we use a three-layer CNN and the CIFAR-10
dataset, and average over three random seeds.

Similarly as in the single-machine case, we find that the equilibrium sharpness is
inversely proportional to the learning rate $\eta$; doubling the learning rate from
$0.005$ to $0.01$ cuts the equilibrium value in half.
\begin{figure*}
    \includegraphics[width=1.0\linewidth]{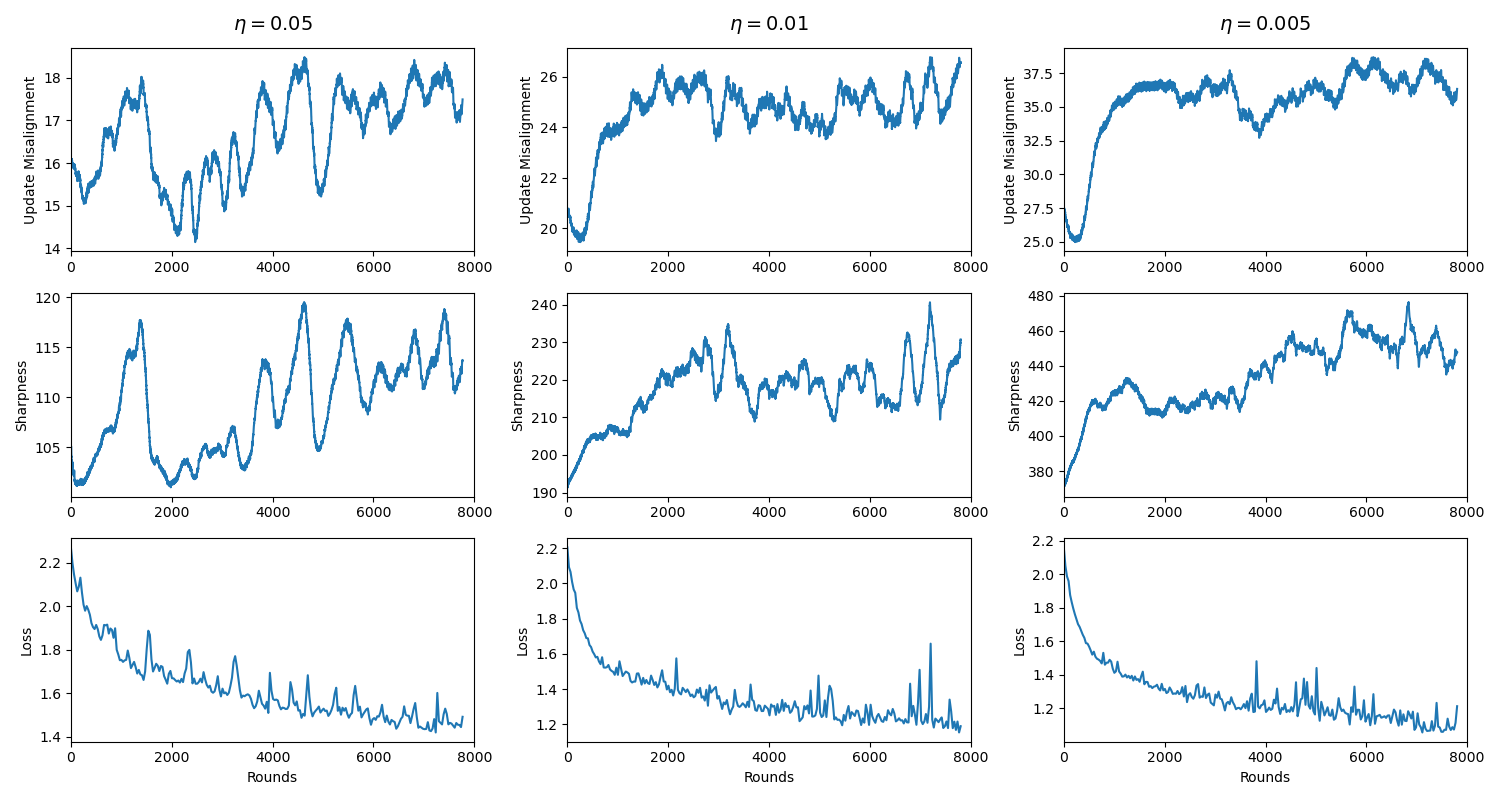}
    \caption{Update misalignment (top), local sharpness (middle), and training loss (bottom) of SCAFFOLD with various learning rates on CIFAR-10 using a shallow CNN architecture. Update misalignment and sharpness are clearly correlated, implying that sharpness causes instability and drift in SCAFFOLD. }
    \label{fig:error-sharp}
\end{figure*}
Interestingly, the data heterogeneity $h$ also appears to influence the equilibrium
sharpness, though less so than the learning rate, and in different ways for FedAvg and
SCAFFOLD. For FedAvg, higher heterogeneity tends to decrease the equilibrium sharpness,
going from about $550$ to about $350$ as heterogeneity goes from $h=0$ to $h=0.9$. For
SCAFFOLD, the equilibrium sharpness is less influenced by heterogeneity, except at
$h=0.9$ for which the behavior is very erratic. Given that the equilibrium value depends
on our hyperparameter $h$ through the opaque procedure of non-i.i.d. data shuffling, we
expect that the equilibrium sharpness value is not expressed by a simple formula in
terms of $h$.


Lastly, for both algorithms the communication interval $K$ has almost no effect on the
equilibrium value, which has
important implications for SCAFFOLD's theory-practice gap. Recall that SCAFFOLD's
theoretical guarantees rely on the condition that the sharpness along the trajectory is
bounded by $\mathcal{O}(1/(\eta K))$, which ensures that SCAFFOLD's estimate of the
global gradient has controllable error. Notice that the threshold $\mathcal{O}(1/(\eta
K))$ vanishes as $K$ becomes large, yet Figure \ref{fig:sharp-char} shows that the
actual sharpness observed along the trajectory is not influenced by $K$. This strongly
suggests that large communication intervals will violate SCAFFOLD's theoretical
justification when EoS occurs. In the following, we will
empirically measure the error in SCAFFOLD's approximation and establish a connection to
EoS.


\subsection{High Sharpness Empirically Correlates with SCAFFOLD Misalignment} \label{sec:exp_misalign}
To quantify how well SCAFFOLD and FedAvg can match the ground truth
global gradient, we introduce a new metric, called \textit{update
misalignment}. At every iteration (communication round $t$, local step $k$), this metric
tracks the error between the local model update $\Delta_i^{t,k} := (w_i^{t,k+1} -
w_i^{t,k}) / \eta$ and the global gradient \textit{evaluated at the local model}:
\begin{equation}
    D^{t,k} = \frac{1}{M} \sum_{i=1}^M \left\| \Delta_i^{t,k} - \nabla f(w_i^{t,k}) \right\|
\end{equation}
Note that $\Delta_i^{t,k} = g_i^{t,k}$ for FedAvg, and $\Delta_i^{t,k} = g_i^{t,k} -
c_i^t + c^t$ for SCAFFOLD. In theory, SCAFFOLD's update misalignment should be small,
while FedAvg's should scale as data heterogeneity worsens.

As seen in Figure \ref{fig:error-sharp}, there is a strong, positive correlation between
update misalignment and sharpness along individual trajectories for multiple learning
rates. For any single trajectory, the two curves look remarkably similar. Table \ref{tab:correlations} shows that the
detrended correlation \citep{Podobnik_2008} between update misalignment $D^{t,k}$ and
sharpness is quite high for SCAFFOLD ($\geq 0.54$) and near zero for FedAvg. We use the
detrended correlation to remove global correlations (e.g. both sharpness and update
misalignment gradually increase through training) and focus on more fine-grained oscillations. In
general, we found that the vanilla correlation between update misalignment and sharpness
is even higher than the detrended correlation.

\begin{figure}[t]
\centering
\includegraphics[width=\columnwidth]{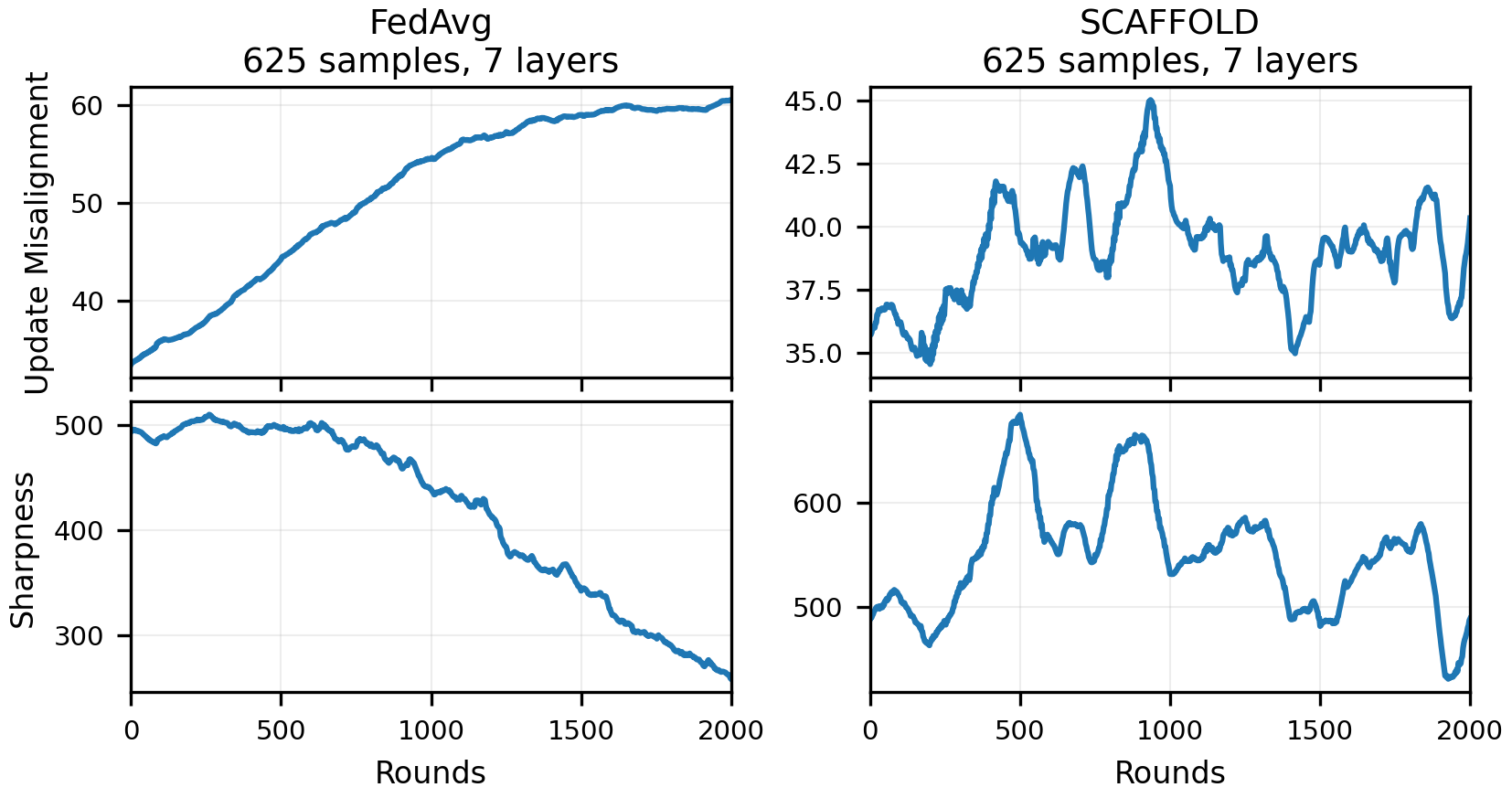}
\caption{Update misalignment and sharpness for FedAvg and SCAFFOLD on CIFAR-10 with 625 samples and 7 layers. The peaks and valleys of SCAFFOLD's update misalignment closely track that of the sharpness, unlike FedAvg.}
\label{fig:cifar-fedavg-scaffold-625-7}
\end{figure}

Essentially, this means that SCAFFOLD's error in estimating the global gradient is much
larger for \textit{steps} in the trajectory where the sharpness is high, particularly at the
EoS. On the contrary, FedAvg's update error is unaffected by high sharpness. This
phenomenon is very consistent across learning rates (Figure \ref{fig:error-sharp}) and
network depth/dataset size (Table \ref{tab:correlations}, Figure
\ref{fig:cifar-fedavg-scaffold-625-7}). These results suggest a \textit{mechanism} for
SCAFFOLD's inability to outperform FedAvg in deep learning: \textbf{when the sharpness
is near the EoS threshold, SCAFFOLD cannot estimate the global gradient}.

\begin{table}[t]
\centering
\small

\renewcommand{\arraystretch}{1.14}

\begin{tabular}{@{} c c
    S[table-format=1.3]
    S[table-format=1.3] @{}}
\toprule
\textbf{Dataset Size} & \textbf{CNN Layers} & \textbf{SCAFFOLD} & \textbf{FedAvg} \\
\midrule
1250 & 3 & 0.813 & 0.162 \\
1250 & 5 & 0.650 & 0.031 \\
1250 & 7 & 0.544 & -0.157 \\
\midrule
625  & 3 & 0.628 & 0.045 \\
625  & 5 & 0.579 & -0.014 \\
625  & 7 & 0.611 & 0.128 \\
\bottomrule
\end{tabular}

\caption{Detrended correlation between sharpness and update misalignment for SCAFFOLD and FedAvg (CNN, CIFAR-10). SCAFFOLD's update misalignment is highly correlated with sharpness, while for FedAvg the update misalignment is essentially unrelated to sharpness.}
\label{tab:correlations}
\end{table}



\begin{table}[t]
\centering

\resizebox{\columnwidth}{!}{%
\begin{tabular}{@{}c c c c c c@{}}

\toprule
& & \multicolumn{2}{c}{\textbf{CIFAR}} & \multicolumn{2}{c}{\textbf{MNIST}} \\
\cmidrule(r){3-4} \cmidrule(l){5-6}
\textbf{Size} & \textbf{Layers}
& \textbf{FedAvg} & \textbf{SCAF.}
& \textbf{FedAvg} & \textbf{SCAF.} \\
\midrule
1250 & 3 & \textbf{0.870} & 0.960 & 0.0642 & \textbf{0.0067} \\
1250 & 5 & \textbf{0.790} & 0.850 & 0.0749 & \textbf{0.0092} \\
1250 & 7 & \textbf{0.820} & 0.840 & 0.0489 & \textbf{0.0062} \\
\midrule
625 & 3 & \textbf{0.950} & 1.050 & 0.0612 & \textbf{0.0019} \\
625 & 5 & \textbf{0.700} & 0.830 & 0.0700 & \textbf{0.0009} \\
625 & 7 & \textbf{0.750} & 0.910 & 0.0276 & \textbf{0.0002} \\
\bottomrule
\end{tabular}
}
\caption{Training loss of FedAvg and SCAFFOLD across dataset sizes and network depth for CIFAR-10 and MNIST. For MNIST, SCAFFOLD consistently outperforms FedAvg, while for CIFAR the reverse is true. Learning rates, between 0.05, 0.01, and 0.005 were tuned separately for each algorithm, dataset size, and network depth. FedAvg consistently selected $\eta=0.05$, while SCAFFOLD selected $\eta=0.01$.}

\label{tab:cifar_mnist_losses}
\end{table}

\begin{figure}[t]
\centering
\includegraphics[height=1.5in]{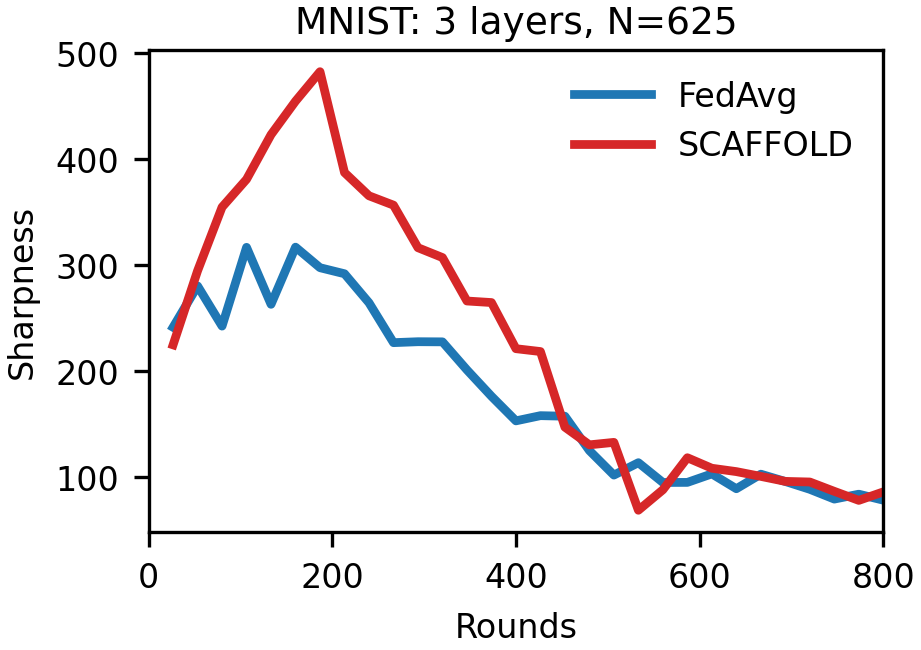}
\caption{Sharpness of the local objective during MNIST training. Neither algorithm exhibits progressive sharpening.}
\label{fig:mnist_sharp}
\end{figure}

Importantly, high update misalignment under EoS translates to actual decline in training
performance, as verified by the fact that FedAvg attains lower loss than SCAFFOLD for
settings with EoS, but not in settings without. This is shown in Table
\ref{tab:cifar_mnist_losses}, where we compare FedAvg and SCAFFOLD for both CIFAR-10 and
MNIST, under various dataset sizes and network depths. We can see that for CIFAR-10,
where EoS actually occurs (see Figures \ref{fig:cif-sharp}, \ref{fig:error-sharp}),
FedAvg and SCAFFOLD are comparable, with FedAvg achieving slightly lower losses. The
MNIST dataset is a different story. In Figure \ref{fig:mnist_sharp}, notice that sharpness quickly decreases towards zero
for MNIST, and in this case SCAFFOLD is achieving near zero loss, far less than FedAvg.
Therefore, SCAFFOLD can outperform FedAvg when the dataset is easy enough that EoS does not occur.

This example provides an important clarification about SCAFFOLD's performance that is
not covered by previous explanations. For example, \citet{yu2022tct} argue that SCAFFOLD
struggles under the nonconvexity induced by deep architectures, and propose a fix based
on eliminating nonconvexity. From Table \ref{tab:cifar_mnist_losses}, we see that
SCAFFOLD can perform its intended function perfectly well for networks up to seven
layers, as long as sharpness stays low, despite the nonconvexity. Rather than the deep
architecture, it is the dataset complexity that indirectly hinders SCAFFOLD through the
presence of high sharpess at the EoS.

\section{Conclusion} \label{sec:conclusion}

We propose that progressive sharpening and EoS act as a mechanism that causes SCAFFOLD to underperform in federated optimization of deep networks.
First, we establish that both FedAvg and SCAFFOLD exhibit progressive sharpening and EoS in federated settings.
We also measure the equilibrium value of the sharpness along the trajectory, and find it is inversely proportional to the learning rate (similarly as GD), while exhibiting a smaller dependence on the data heterogeneity, and negligible dependence on the number of local steps.
This shows that, when the number of local steps is large and the sharpness hovers around its equilibrium, SCAFFOLD cannot satisfy its theoretical requirement that sharpness along the trajectory be inversely proportional to the number of local steps (see Section \ref{sec:preliminaries}).
Lastly, we show that SCAFFOLD's error in estimating the global gradient (quantified by update misalignment) is strongly correlated with sharpness along the trajectory; accordingly, SCAFFOLD fails to estimate the global gradient at the EoS.

Contrary to previous work that investigates SCAFFOLD's underperformance \citep{yu2022tct}, we observe in our experiments that SCAFFOLD can work well in deep learning in some situations, particularly those in which EoS does not occur.
Accordingly, we posit that SCAFFOLD's underperformance is not due to the nonconvexity of the loss landscape induced by deep architectures, but rather the optimization dynamics peculiar to deep learning.

\textbf{Limitations}
In this paper we have mainly focused on versions of FedAvg and SCAFFOLD that use full-batch gradients (as in most works on EoS \citep{cohen2021gradient, cohen2023adaptive, cohen2025understanding}), for training CNNs for image classification tasks.
These settings can naturally be extended to include stochastic gradients, other network architectures such as Transformers, larger image datasets, and language tasks.
Another limitation of our work is the fact that we do not determine the equilibrium sharpness value in closed-form for either FedAvg or SCAFFOLD.
Given that the sharpness seems to depend on data heterogeneity, and we induce data heterogeneity through an ad hoc procedure with parameter $h$ controlling the degree of heterogeneity, we do not expect the sharpness equilibrium value to have a simple closed form in terms of $h$. Rather, a closed-form equilibrium value should likely incorporate data heterogeneity in terms of properties of the loss landscape, rather than the parameter $h$ itself.

\clearpage

\section*{Acknowledgements}
Computations were run on the Hopper cluster provided by the Office of Research Computing at George Mason University. Anant Khandelwal acknowledges the support of the Aspiring Scientists Summer Internship Program. At the time of completing this work, Michael Crawshaw was supported by the Doctoral Research Scholarship of George Mason University. Mingrui Liu is supported by NSF grants \#2436217, \#2425687, \#2601681.

\bibliography{aaai2027}



\clearpage

\appendix
\section{Experimental Details}

\subsection{Neural Network Architectures}
\label{app:model-architectures}

All networks produce ten unnormalized output logits. No softmax or other
activation is applied after the final classification layer. Unless stated
otherwise, all convolutional and fully connected layers include a bias.

We use the rectified linear unit
\[
\operatorname{ReLU}(z)=\max(0,z)
\]
throughout the CNNs and in the CIFAR-10 MLP. ReLU sets negative inputs to
zero and leaves positive inputs unchanged. It is piecewise linear and does
not saturate for positive inputs.

\paragraph{CIFAR-10 CNNs.}

CIFAR-10 images have dimension $3\times32\times32$. In the
variable-depth CNN, the depth $L$ denotes the number of convolutional
layers; the final linear classifier is not included in this count. Each
convolution uses a $3\times3$ kernel, stride one, and padding one. The
first convolution maps the three input channels to 32 feature channels,
and every subsequent convolution preserves 32 channels. A ReLU follows
every convolution.

For $L\ge3$, the architecture consists of repeated two-convolution
blocks,
\[
\operatorname{Conv}
\rightarrow
\operatorname{ReLU}
\rightarrow
\operatorname{Conv}
\rightarrow
\operatorname{ReLU}
\rightarrow
\operatorname{MaxPool},
\]
with one additional Conv--ReLU pair appended when $L$ is odd. The
resulting feature tensor is flattened and mapped directly to ten logits.

For example, the $L=5$ architecture is
\[
\begin{aligned}
3\times32\times32
&\rightarrow
\operatorname{Conv}(3,32)
\rightarrow
\operatorname{ReLU}
\\
&\rightarrow
\operatorname{Conv}(32,32)
\rightarrow
\operatorname{ReLU}
\\
&\rightarrow
\operatorname{MaxPool}
\rightarrow
\operatorname{Conv}(32,32)
\\
&\rightarrow
\operatorname{ReLU}
\rightarrow
\operatorname{Conv}(32,32)
\\
&\rightarrow
\operatorname{ReLU}
\rightarrow
\operatorname{MaxPool}
\\
&\rightarrow
\operatorname{Conv}(32,32)
\rightarrow
\operatorname{ReLU}
\\
&\rightarrow
\operatorname{Flatten}
\rightarrow
\operatorname{Linear}(2048,10).
\end{aligned}
\]

For $L=3$, this architecture is identical to the one used in
\citet{cohen2021gradient}, and it is extended naturally for $L>3$.

\paragraph{MNIST CNNs.}

MNIST images have dimension $1\times28\times28$. The variable-depth
MNIST CNN follows the same overall structure as the CIFAR-10 CNN. The
number of channels increases over the first three convolutions,
\[
1\rightarrow32\rightarrow64\rightarrow128,
\]
after which all remaining convolutions preserve 128 channels. Every
convolution uses a $3\times3$ kernel with stride one and padding one,
followed by ReLU. A $2\times2$ max-pooling layer follows every second
convolution.

The $L=5$ architecture is
\[
\begin{aligned}
1\times28\times28
&\rightarrow
\operatorname{Conv}(1,32)
\rightarrow
\operatorname{ReLU}
\\
&\rightarrow
\operatorname{Conv}(32,64)
\rightarrow
\operatorname{ReLU}
\\
&\rightarrow
\operatorname{MaxPool}
\rightarrow
\operatorname{Conv}(64,128)
\\
&\rightarrow
\operatorname{ReLU}
\rightarrow
\operatorname{Conv}(128,128)
\\
&\rightarrow
\operatorname{ReLU}
\rightarrow
\operatorname{MaxPool}
\\
&\rightarrow
\operatorname{Conv}(128,128)
\rightarrow
\operatorname{ReLU}
\\
&\rightarrow
\operatorname{Flatten}
\rightarrow
\operatorname{Linear}(6272,10).
\end{aligned}
\]

After the two pooling operations, the feature tensor has dimension
$128\times7\times7$, giving
$128\cdot7\cdot7=6272$
inputs to the classifier.

\paragraph{CIFAR-10 MLP.}

The fixed CIFAR-10 MLP first flattens each image into
$3\cdot32\cdot32=3072$ features. It then applies one hidden layer of
width 200 followed by ReLU and a linear classifier:

\[
\begin{aligned}
3072
&\rightarrow
\operatorname{Linear}(3072,200)
\rightarrow
\operatorname{ReLU}
\\
&\rightarrow
\operatorname{Linear}(200,10).
\end{aligned}
\]

This architecture is taken directly from
\citet{cohen2021gradient}.

\paragraph{MNIST MLP.}

The fixed MNIST MLP flattens each image into
$28\cdot28=784$ features and uses two hidden layers of width 1000:

\[
\begin{aligned}
784
&\rightarrow
\operatorname{Linear}(784,1000)
\rightarrow
\operatorname{ReLU}
\\
&\rightarrow
\operatorname{Linear}(1000,1000)
\rightarrow
\operatorname{ReLU}
\\
&\rightarrow
\operatorname{Linear}(1000,10).
\end{aligned}
\]

The final layer produces the ten class logits without an output
activation.

\subsection{Global and Local Sharpness} \label{app:sharp}

For an objective $g$ evaluated at model parameters $w$, we define its
sharpness as the largest eigenvalue of its Hessian:
\[
\operatorname{Sharp}(g,w)
=
\lambda_{\max}\!\left(\nabla^2 g(w)\right).
\]

We define the \textit{global sharpness} at the end of round $t$ as the
sharpness of this averaged model on the global objective:
\[
\mathcal{S}_{\mathrm{global}}^{t}
=
\lambda_{\max}\!\left(
\nabla^2 f\!\left(w^{t+1}\right)
\right),
\qquad
f(w)=\frac{1}{M}\sum_{i=1}^{M}f_i(w).
\]
Thus, global sharpness measures the curvature of the combined training
loss at the model produced by federated averaging. The \textit{local sharpness} at local step $k$ of round $t$ is the average sharpness of the participating clients' local objectives, evaluated at their respective local models:
\[
\mathcal{S}_{\mathrm{local}}^{t,k}
=
\frac{1}{|\mathcal{S}_t|}
\sum_{i\in\mathcal{S}_t}
\lambda_{\max}\!\left(
\nabla^2 f_i\!\left(w_i^{t,k}\right)
\right).
\]
Unlike global sharpness, local sharpness follows the curvature encountered
by the individual client models as they move away from the shared model
during local training. In implementation, the leading eigenvalue is
estimated using the Lanczos method, with Hessian--vector products computed
by automatic differentiation.

\subsection{Heterogeneity Protocol} \label{app:het}
Similar to prior work \citet{karimireddy2020scaffold}, including SCAFFOLD, we simulate client-level statistical heterogeneity using a label-skew partitioning scheme controlled by a parameter \(h \in [0,1]\), where \(h=0\) corresponds to fully i.i.d.\ data and larger values induce greater non-i.i.d.\ structure. After forming the training split, we group examples by class label. For each class, a fraction \(1-h\) of examples is assigned to a global i.i.d.\ pool and the remaining fraction \(h\) is assigned to a global non-i.i.d.\ pool. The i.i.d.\ pool is shuffled, while the non-i.i.d.\ pool is left ordered by label, so that contiguous chunks are label-concentrated.

We then partition both pools evenly across the \(N\) clients and give each client one shard from the i.i.d.\ pool and one shard from the non-i.i.d.\ pool. Thus, every client receives the same proportion of shared and label-skewed data, with \(h\) directly controlling the severity of the skew. When \(h=0\), all client datasets have i.i.d. labels; as \(h\) increases, clients become increasingly specialized to different subsets of classes. For datasets with predefined users (e.g., Sent140), we apply the same idea at the user level by sorting users in the non-i.i.d.\ pool by label proportion before partitioning.

\subsection{Algorithm Pseudocode}

\begin{algorithm}[t]
\caption{\textbf{FedAvg}: Federated Averaging}
\label{alg:fedavg}
\small
\begin{algorithmic}[1]
\Require Initial global model $w^0$, learning rate $\eta$, communication rounds $T$, local steps $K$
\For{$t = 0,1,\dots,T-1$}
    \For{client $i = 1, \ldots, M$ \textbf{in parallel}}
        \State $w_i^{t,0} \gets w^t$
        \For{local step $k = 0,1,\dots,K-1$}
            \State $w_i^{t,k+1} \gets w_i^{t,k} - \eta \nabla f_i(w_i^{t,k})$
        \EndFor
        \State $w_i^{t+1} \gets w_i^{t,K}$
        \State Send $\Delta w_i^t = w_i^{t+1} - w^t$ to server
    \EndFor
    \State $w^{t+1} \gets w^t + \frac{1}{M}\sum_{i=1}^M \Delta w_i^t$
\EndFor
\end{algorithmic}
\end{algorithm}

As in Algorithm \ref{alg:fedavg}, we use FedAvg as a baseline that runs gradient descent on many different datasets and regularly averages models according to a set communication interval. 

\begin{algorithm}[t]
\caption{\textbf{SCAFFOLD}: Stochastic Controlled Averaging for Federated Learning}
\label{alg:scaffold}
\small
\begin{algorithmic}[1]
\Require Initial global model $w^0$, server control variate $c^0$, client control variates $\{c_i^0\}_{i=1}^N$, learning rate $\eta$, communication rounds $T$, local steps $K$
\For{$t = 0,1,\dots,T-1$}
    \For{client $i = 1, \ldots, M$ \textbf{in parallel}}
        \State $w_i^{t,0} \gets w^t$
        \For{local step $k = 0,1,\dots,K-1$}
            \State $w_i^{t,k+1} \gets w_i^{t,k} - \eta\left(\nabla f_i(w_i^{t,k}) - c_i^t + c^t\right)$
        \EndFor
        \State $w_i^{t+1} \gets w_i^{t,K}$
        \State $c_i^{t+1} \gets c_i^t - c^t + \frac{1}{K\eta}(w^t - w_i^{t+1})$
        \State Send $\Delta w_i^t = w_i^{t+1} - w^t$ and $\Delta c_i^t = c_i^{t+1} - c_i^t$ to server
    \EndFor
    \State $w^{t+1} \gets w^t + \frac{1}{M}\sum_{i=1}^M \Delta w_i^t$
    \State $c^{t+1} \gets c^t + \frac{1}{M}\sum_{i=1}^M \Delta c_i^t$
\EndFor
\end{algorithmic}
\end{algorithm}

SCAFFOLD, implemented as in Algorithm \ref{alg:scaffold}, adds control variates to the FedAvg algorithm to reduce heterogeneity. By subtracting previous local gradients and adding previous global gradients, SCAFFOLD adjusts model updates toward ground-truth global gradients. 
\vfil\penalty-10000
\section{Additional Experimental Results} \label{app:extra_exp}

\begin{figure*}
\includegraphics[width=1.0\linewidth]{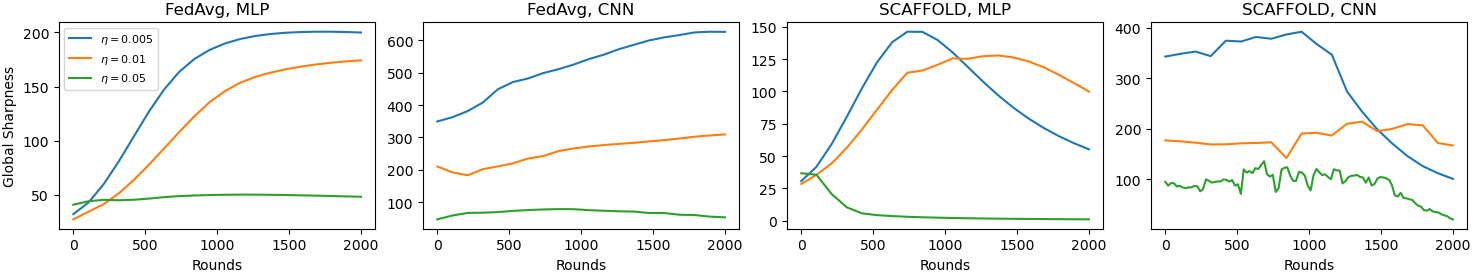}
\includegraphics[width=1.0\linewidth]{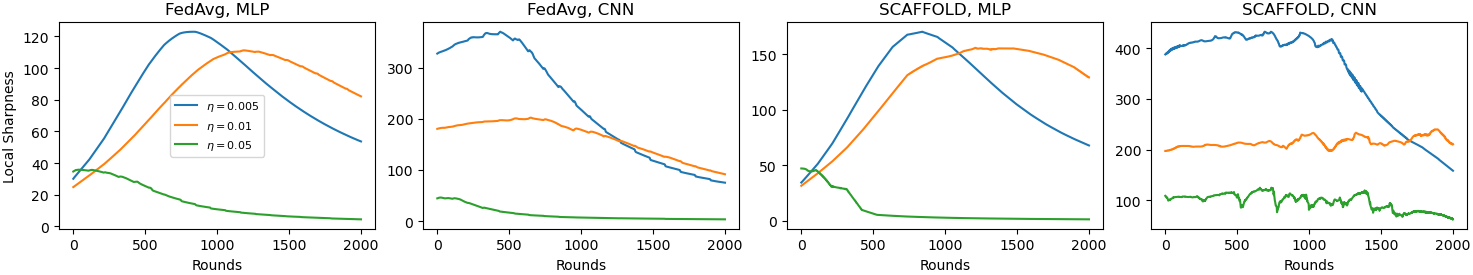}
\includegraphics[width=1.0\linewidth]{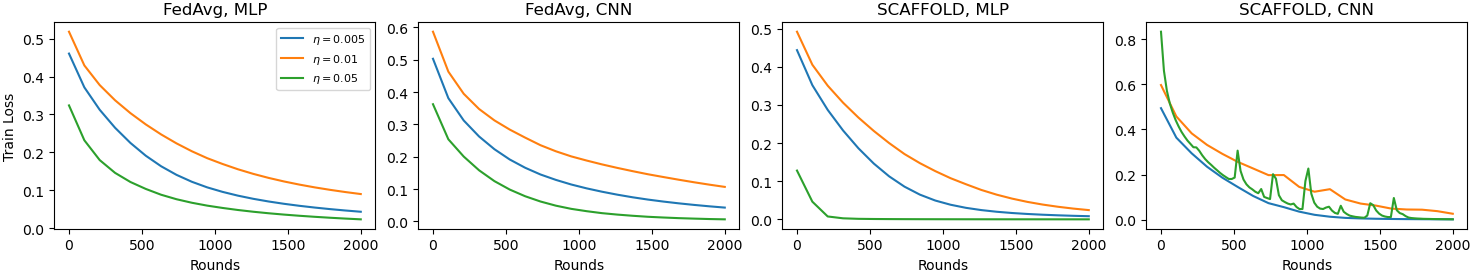}
\caption{FashionMNIST global sharpness (top), local sharpness (middle), and train loss (bottom) over train trajectories with both SCAFFOLD and FedAvg and with both MLP and CNN architectures.}
\label{fig:fashionmnist}
\end{figure*}

Federated learning algorithms on FashionMNIST (see Figure \ref{fig:fashionmnist}) see expected and consistent behavior for sharpness given the relative simplicity of the dataset. Instead of continuous incrase in sharpness until a threshold, both FedAvg and SCAFFOLD experience a modest increase followed by a rapid decrease in sharpness as the model begins to converge and overfit with the maximum margin bias causing smoother optimization. In some cases, sharpness does remain at a threshold, and in other cases sharpness never rises. However, there is still a strong inverse correlation between sharpness and learning rate. 

\begin{figure*}
    \includegraphics[width=1.0\linewidth]{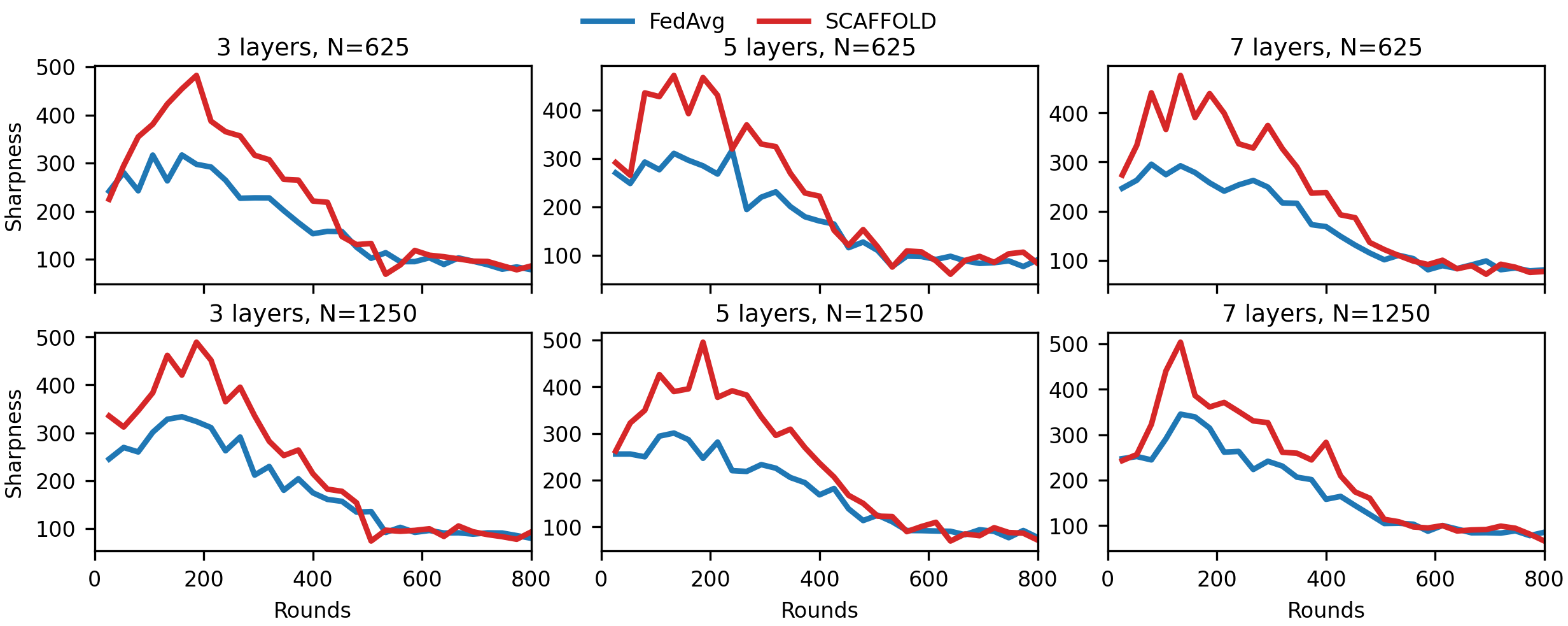}
    \caption{MNIST local sharpness values during training for many CNN layer depths and dataset sizes, as referenced in \ref{tab:cifar_mnist_losses}. For MNIST, sharpness rises initially, and then gradually decreases as the network converges to 0 loss. }
    \label{fig:mnist-sharp}
\end{figure*}

For MNIST experiments, shown in Figure \ref{fig:mnist-sharp}, an even simpler dataset than FashionMNIST, sharpness consistently rises modestly followed by a gradual decline to small values as any architecture begins to fit the data close to perfectly. This indicates that our results, which have high levels of SCAFFOLD performance relative to FedAvg on MNIST tasks, may show that SCAFFOLD suffers from high sharpness values. These training runs are identical to those shown in Table \ref{tab:cifar_mnist_losses} for MNIST. These results, paired with those showing SCAFFOLD outperforming FedAvg for MNIST, support our conjecture that sharpness, not convexity, explains SCAFFOLD failures.

\begin{figure*}[t]

        \includegraphics[width=\linewidth]{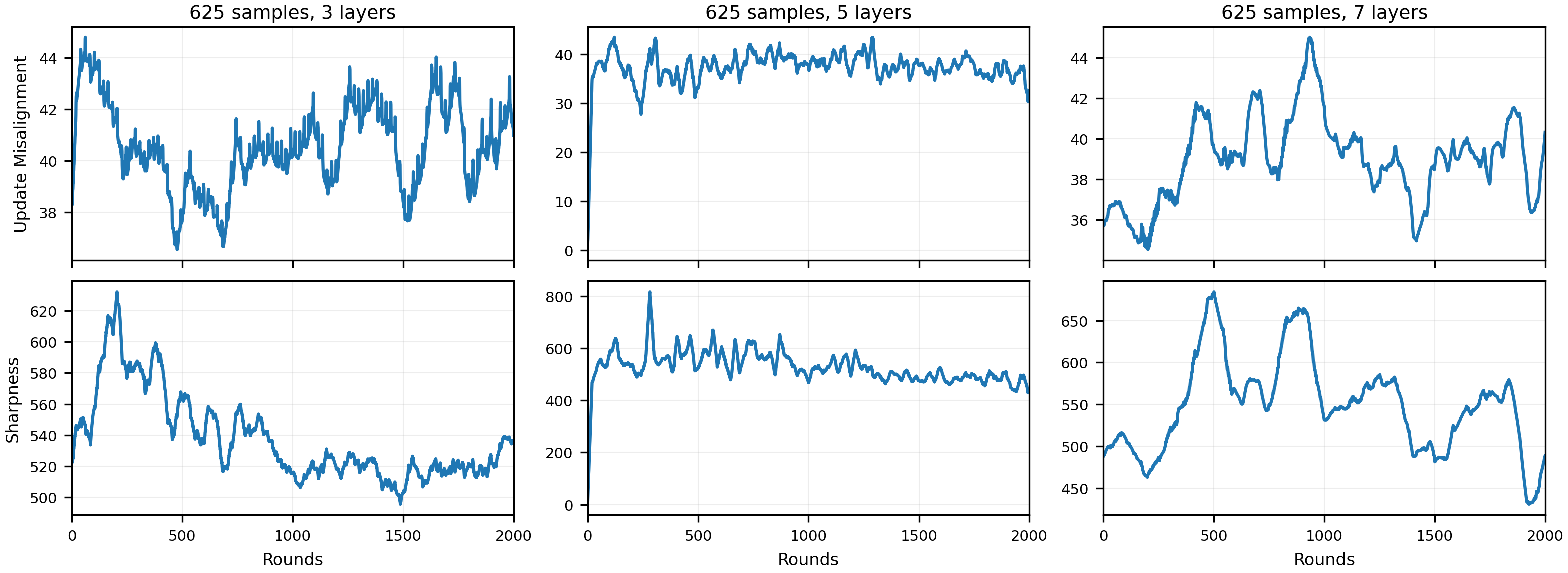}
      
        \includegraphics[width=\linewidth]{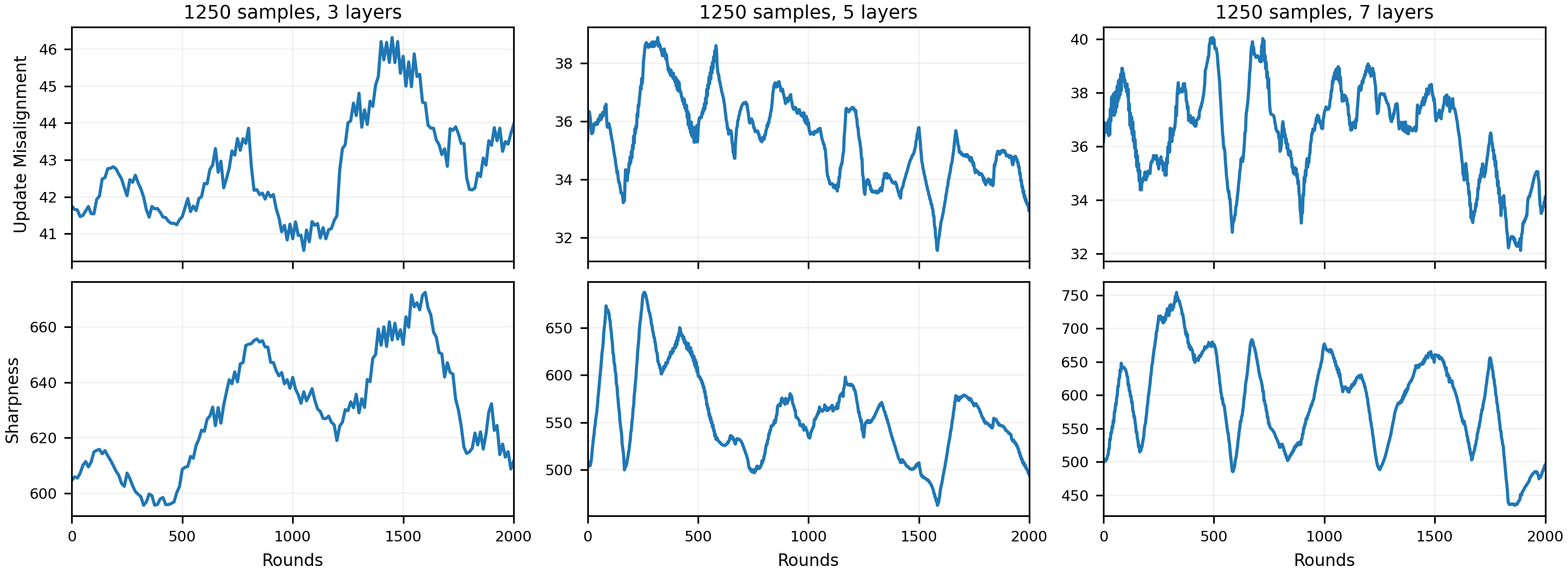}

    \caption{Update misalignment and local sharpness values over SCAFFOLD trajectories for Table \ref{tab:cifar_mnist_losses}. Update misalignment and sharpness stay highly correlated across different architectures and dataset sizes for SCAFFOLD. }
    \label{fig:appendix-extra-vals}
\end{figure*}

In Table \ref{tab:cifar_mnist_losses} trajectories we consistently observe similar correlations between the local sharpness and the update misalignment. These have high (>0.5) degrees of correlation between normalized values across model depths and dataset sizes, demonstrating the effect of sharpness on misalignment. Qualitatively, trajectories in Figure \ref{fig:appendix-extra-vals} clearly show a match between the oscillations present, indicating major and observable differences.

\begin{figure*}[t]
\includegraphics[width=\linewidth]{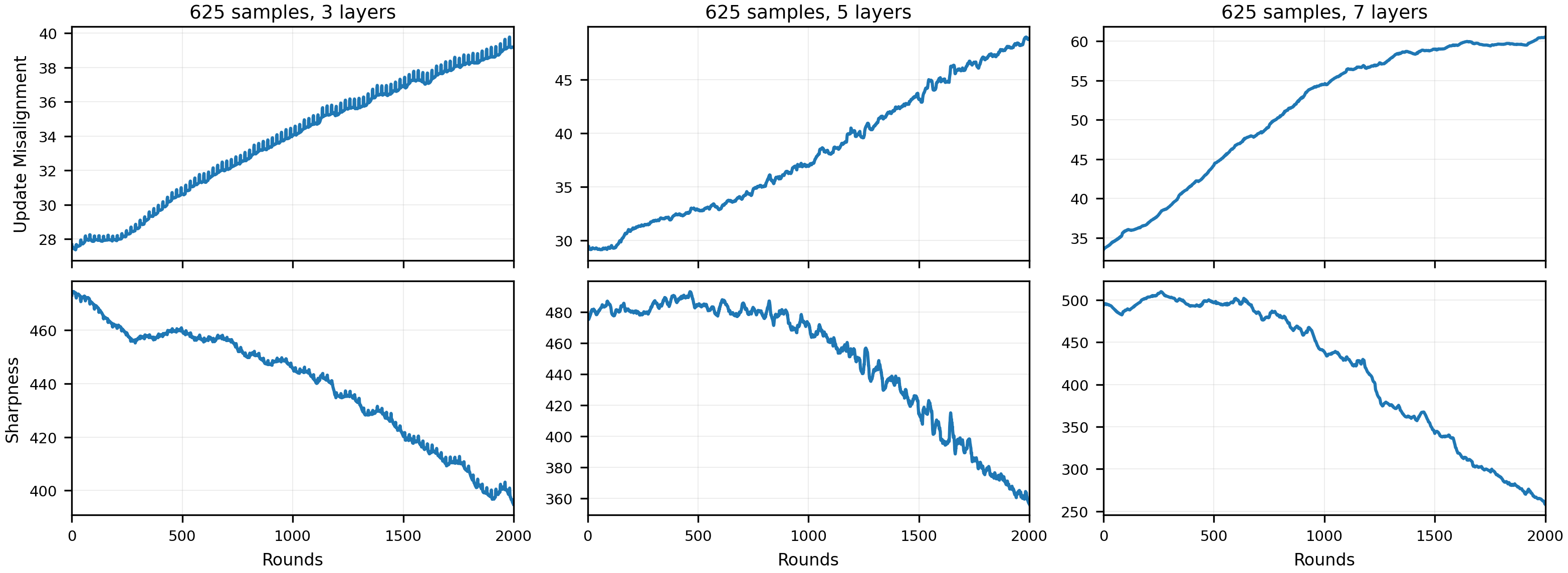}
      
        \includegraphics[width=\linewidth]{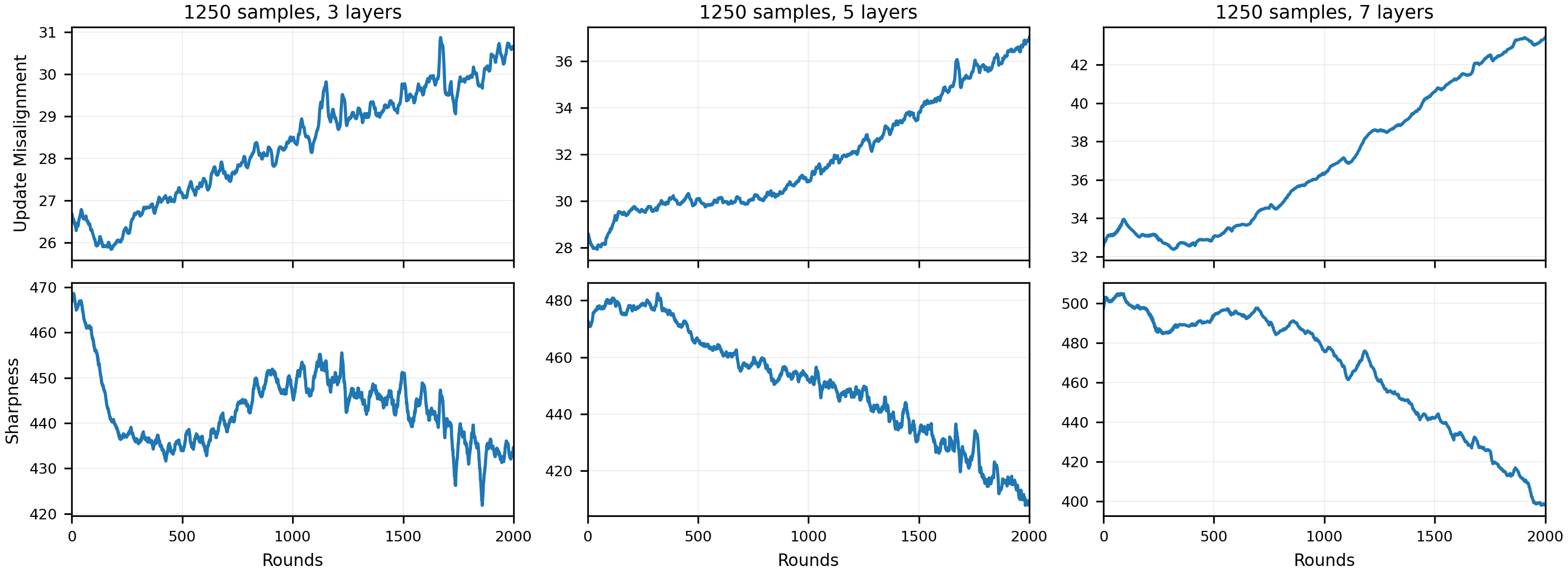}

    \caption{Update misalignment and local sharpness values over FedAvg trajectories for Table \ref{tab:cifar_mnist_losses}.  For FedAvg, update misalignment and sharpness are empirically disassociated.}
    \label{fig:appendix-extra-vals2}
\end{figure*}

In comparison to SCAFFOLD, FedAvg trajectories from Table \ref{tab:cifar_mnist_losses} shows little or even negative correlation between update misalignment and SCAFFOLD. Figure \ref{fig:appendix-extra-vals2} shows no qualitative similarity between the curves across different dataset sizes and model depths. There are less volatile movements in each metric, yielding no particular similarities in oscillations, and throughout training, these two metrics have opposite directions of growth. Note that sharpness here decreases throughout training due to high sharpness at initialization--starting weights and datasets yielded sharpness values > 500, causing sharpness to stagnate or decrease in training. 

\begin{figure*}
    \centering
    \includegraphics[width=\linewidth]{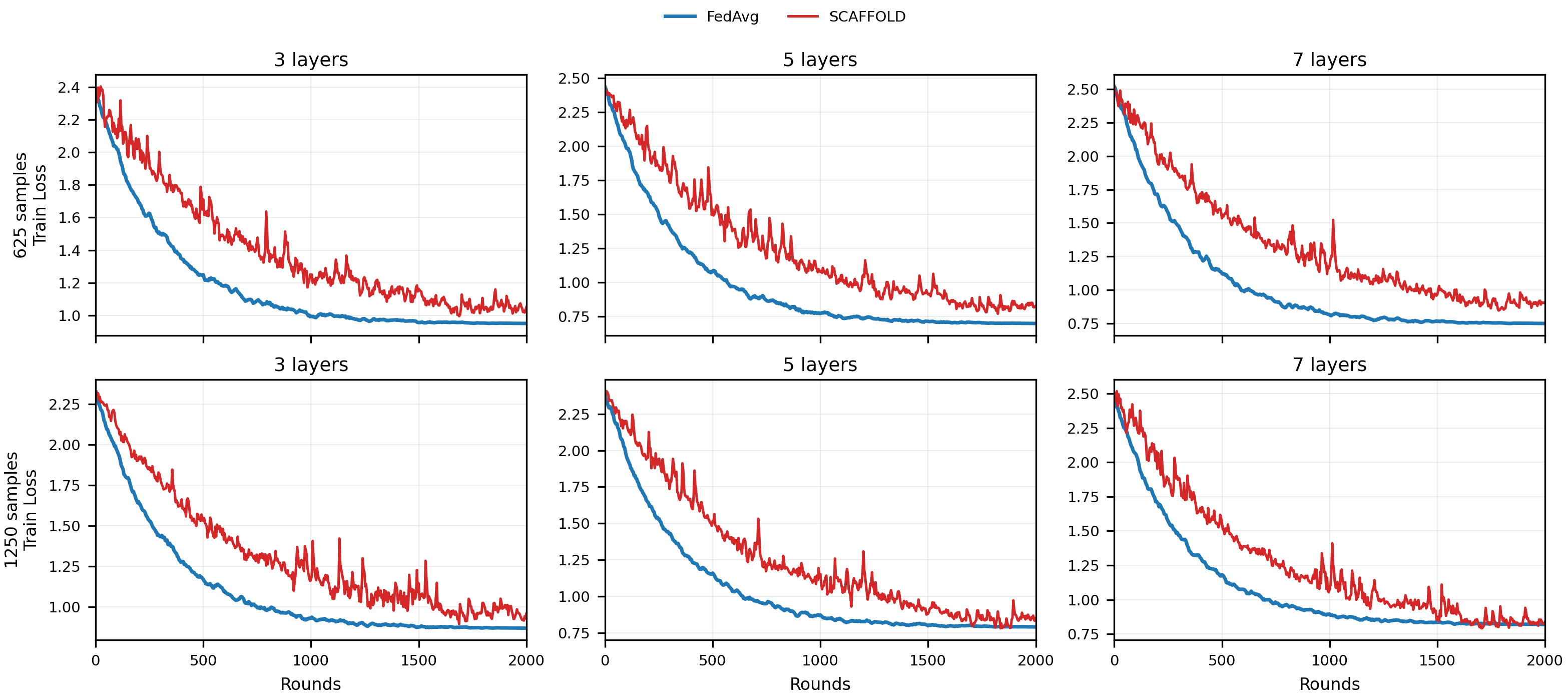}
    \caption{Loss curves for \ref{tab:cifar_mnist_losses} with a CIFAR dataset for both SCAFFOLD and FedAvg for CNNs of depth 3, 5, and 7 and dataset sizes of 625 and 1250. FedAvg consistently attains lower loss and converges faster, with less instability in loss. }
    \label{fig:cifar-loss-curves}
\end{figure*}

Figure \ref{fig:cifar-loss-curves} shows loss trajectories for both SCAFFOLD and FedAvg under the settings from \ref{tab:cifar_mnist_losses}. SCAFFOLD takes longer to converge, with significantly more instability along its path. This reflects that not only does FedAvg outperform on final loss, but it is also easier and more consistent to train. Paired with our results on update misalignment, this suggests that high sharpness causes SCAFFOLD updates to be less accurate and less stable.
\end{document}